\documentclass{article}

\usepackage[numbers]{natbib}

\usepackage[final]{neurips_2025}

\usepackage[utf8]{inputenc} 
\usepackage{float}
\usepackage[T1]{fontenc}    
\usepackage{hyperref}       
\usepackage{url}            
\usepackage{booktabs}       
\usepackage{amsfonts}       
\usepackage{nicefrac}       
\usepackage{microtype}      
\usepackage{xcolor}         
\usepackage{amsmath}
\usepackage[linesnumbered,ruled,vlined]{algorithm2e}
\usepackage{graphicx}
\usepackage{amsthm}
\usepackage{subcaption}

\title{Quantum Graph Attention Network: A Novel Quantum Multi-Head Attention Mechanism for Graph Learning}

\author{%
  An Ning \\
  School of Electrical Engineering \\
  Korea Advanced Institute of Science and Technology (KAIST)\\
  \texttt{ansonning@kaist.ac.kr} \\
  \AND
   Tai Yue Li, Nan Yow Chen \\
  National Center for High-performance Computing (NCHC)\\
}

\begin{document}

\maketitle

\begin{abstract}

We propose the Quantum Graph Attention Network (QGAT), a hybrid graph neural network that integrates variational quantum circuits into the attention mechanism. At its core, QGAT employs strongly entangling quantum circuits with amplitude-encoded node features to enable expressive nonlinear interactions. Distinct from classical multi-head attention that separately computes each head, QGAT leverages a single quantum circuit to simultaneously generate multiple attention coefficients. This quantum parallelism facilitates parameter sharing across heads, substantially reducing computational overhead and model complexity. Classical projection weights and quantum circuit parameters are optimized jointly in an end-to-end manner, ensuring flexible adaptation to learning tasks. Empirical results demonstrate QGAT’s effectiveness in capturing complex structural dependencies and improved generalization in inductive scenarios, highlighting its potential for scalable quantum-enhanced learning across domains such as chemistry, biology, and network analysis. Furthermore, experiments confirm that quantum embedding enhances robustness against feature and structural noise, suggesting advantages in handling real-world noisy data. The modularity of QGAT also ensures straightforward integration into existing architectures, allowing it to easily augment classical attention-based models.

\end{abstract}

\section{Introduction}
Graph Neural Networks (GNNs) have emerged as an effective framework for learning from graph-structured data, demonstrating wide applicability in social networks, recommendation systems, bioinformatics, and molecular property prediction\cite{wu2020comprehensive, scarselli2008graph, zhou2020graph}. Numerous architectures have been developed to model node dependencies, including Graph Convolutional Networks (GCN)\cite{kipf2016semi, wu2019simplifying}, GraphSAGE\cite{hamilton2017inductive, liu2020graphsage}, Graph Isomorphism Networks (GIN)\cite{xu2018powerful}, Graph Attention Networks (GAT)\cite{velivckovic2017graph, wang2019heterogeneous}, GATv2\cite{brody2021attentive}, and Graph Transformers\cite{yun2019graph, kreuzer2021rethinking, rampavsek2022recipe}. These models typically utilize neighborhood aggregation or attention mechanisms to derive node representations. However, classical GNNs often face representational bottlenecks and scalability issues on large or structurally complex graphs, particularly those with intricate relational patterns or nonlinear interactions\cite{xu2018powerful, alon2020bottleneck, oono2019graph}. Attention-based models like GAT and GATv2\cite{velivckovic2017graph, brody2021attentive, dwivedi2022long} partially address these limitations but still struggle to fully capture dynamic nonlinear interactions within complex graph topologies. Hence, developing more expressive and adaptable architectures is necessary to better exploit the richness inherent in real-world graphs\cite{dwivedi2022long}.

Recent advancements in quantum computing suggest that quantum machine learning (QML) models may surpass classical methods in capturing complex nonlinear relationships due to fundamental quantum properties, such as entanglement and superposition. These quantum phenomena enable inherently richer representations within high-dimensional Hilbert spaces\cite{schuld2015introduction, biamonte2017quantum}. Unlike classical neural networks that explicitly introduce nonlinearity through activation functions, quantum models naturally achieve nonlinearity via the unitary evolution of quantum states\cite{schuld2021effect, lloyd2020quantum}. For instance, recent studies on quantum pointwise convolution\cite{ning2024quantum} have shown that quantum circuits encode nonlinearities flexibly and scalably, improving feature representations without classical nonlinear activations. Thus, quantum computing holds significant promise for enhancing the expressive capabilities of machine learning models, especially in tasks involving complex data structures challenging traditional approaches.

Quantum attention mechanisms, particularly quantum self-attention, have been demonstrated as theoretically sound and practically feasible on near-term quantum devices. Recent quantum attention models—such as QSAN\cite{zhang2022qsan}, QKSAN\cite{li2023qksan}, and quantum self-attention networks for text classification\cite{lu2022quantum}—successfully replicate classical self-attention behavior through quantum interference and kernel methods. These models effectively capture long-range dependencies with fewer parameters and improved robustness against overfitting. Furthermore, quantum extensions of Transformer architectures like Quixer\cite{sarai2023quixer}, Quantum Vision Transformers\cite{jiang2023quantumvit}, and QClusformer\cite{shen2023qclusformer} have shown quantum-enhanced performance across various domains, including image classification, text modeling, and clustering. These findings indicate that quantum attention mechanisms are viable and can outperform classical counterparts in modeling complex data structures, underscoring their potential for integration into graph neural networks.

Motivated by these insights, we propose the Quantum Graph Attention Network (QGAT), a hybrid architecture integrating parameterized quantum circuits within graph attention computations. QGAT aims to enhance the expressive power and robustness of GNNs while preserving the scalability and modularity found in classical models like GAT. Specifically, node features are embedded into quantum states using amplitude encoding, and a single quantum circuit generates multiple attention heads simultaneously. This quantum-inspired design reduces the total parameter count, fosters stronger feature entanglement, and better captures complex relational patterns. Additionally, the intrinsic quantum state distributions contribute to increased robustness against noise, positioning QGAT as an effective and scalable alternative to classical attention-based GNNs.

\section{Related Work}
\paragraph{Graph Attention Network.}
Graph Attention Networks (GAT)~\cite{velivckovic2017graph} introduce a learnable self-attention mechanism into graph neural networks, allowing each node to assign different importance weights to its neighbors during message aggregation. Unlike earlier methods that use fixed or uniform weighting schemes, GAT dynamically computes attention coefficients based on node feature similarity. This enables the model to focus on more relevant neighbors, improving adaptability to heterogeneous graph structures while preserving scalability and parallelizability.

\paragraph{GATv2.}
GATv2~\cite{brody2021attentive} improves upon the original GAT by computing attention scores from a joint transformation of both source and target node features before applying the nonlinearity. This change eliminates constraints in the original GAT's expressiveness, enabling it to capture more complex and asymmetric node relationships. GATv2 retains the computational advantages of GAT while providing strictly greater representational power.

\paragraph{Quantum Attention.}
Quantum attention mechanisms aim to enhance classical attention modules by leveraging the high-dimensional, entangled representations offered by quantum systems. Recent studies have shown that variational quantum circuits can function as learnable attention score generators. For example, models such as QSAN~\cite{zhang2022qsan} and QKSAN~\cite{li2023qksan} use parameterized quantum circuits to produce attention weights, where measured expectation values directly correspond to attention coefficients. These quantum-derived scores inherently capture nonlinear dependencies with fewer parameters. Applications in text classification~\cite{lu2022quantum} further demonstrate that quantum self-attention can effectively model semantic relevance, achieving performance on par with or better than classical approaches. These findings support the use of quantum outputs as viable and efficient attention mechanisms.

\section{Methods}

\subsection{Preliminaries}

\paragraph{Graph Attention Network.}

Graph Attention Networks (GAT)~\cite{velivckovic2017graph} leverage masked self-attention mechanisms to aggregate information from neighboring nodes. Specifically, the attention coefficient $\alpha_{ij}$ between node $i$ and node $j$ is computed as:
\begin{equation}
\alpha_{ij} = \frac{\exp\left(\text{LeakyReLU}\left(\mathbf{a}^{\top}[\mathbf{W}\mathbf{h}_i \| \mathbf{W}\mathbf{h}_j]\right)\right)}{\sum\limits_{k \in \mathcal{N}_i} \exp\left(\text{LeakyReLU}\left(\mathbf{a}^{\top}[\mathbf{W}\mathbf{h}_i \| \mathbf{W}\mathbf{h}_k]\right)\right)},
\end{equation}
where $\mathbf{W} \in \mathbb{R}^{d' \times d}$ is a shared weight matrix applied to input features, $\mathbf{a} \in \mathbb{R}^{2d'}$ is a learnable attention vector, $\mathbf{h}_i, \mathbf{h}_j \in \mathbb{R}^d$ are input node features, $\mathcal{N}_i$ denotes the neighbors of node $i$, and $[\cdot \| \cdot]$ indicates feature-wise concatenation. This formulation allows each node to assign different attention weights to its neighbors based on their content, thereby enhancing representational flexibility.

\paragraph{GATv2.}

GATv2~\cite{brody2021attentive} improves upon the original GAT by allowing the attention mechanism to be input-dependent and more expressive. Instead of computing attention on linearly transformed inputs independently, it applies a joint transformation before nonlinearity:
\begin{equation}
\alpha_{ij} = \frac{\exp\left(\mathbf{a}^{\top} \cdot \text{LeakyReLU}(\mathbf{W}[\mathbf{h}_i \| \mathbf{h}_j])\right)}{\sum\limits_{k \in \mathcal{N}_i} \exp\left(\mathbf{a}^{\top} \cdot \text{LeakyReLU}(\mathbf{W}[\mathbf{h}_i \| \mathbf{h}_k])\right)},
\end{equation}
where the symbols retain the same meaning as above. By applying the nonlinearity after the joint linear transformation of both source and target node features, GATv2 eliminates the constraint of static attention patterns and enables asymmetric relational modeling between nodes. This makes GATv2 strictly more expressive while maintaining the efficiency and parallelism of the original design.

\paragraph{Quantum Machine Learning.}
Quantum Machine Learning (QML) integrates quantum computation into machine learning pipelines by leveraging quantum circuits to process and transform data. A core step in QML is \textit{quantum data embedding}, which encodes classical input vectors into quantum states suitable for further quantum processing. Two widely used strategies are angle embedding and amplitude embedding.

In \textit{angle embedding}, each classical feature is mapped to a rotation gate acting on a corresponding qubit:
\begin{equation}
|\psi(\mathbf{x})\rangle = \bigotimes_{i=1}^n R_Y(x_i) |0\rangle,
\end{equation}
where $\mathbf{x} = (x_1, x_2, \ldots, x_n)$ is the input vector and $R_Y(x_i) = \exp(-i x_i Y/2)$ denotes a rotation around the Y-axis on the $i$-th qubit. This method creates shallow, hardware-friendly circuits for encoding real-valued features.

In contrast, \textit{amplitude embedding} directly maps the components of $\mathbf{x}$ into the amplitude vector of a quantum state:
\begin{equation}
|\psi(\mathbf{x})\rangle = \sum_{i=0}^{2^n - 1} x_i |i\rangle,
\end{equation}
where $\mathbf{x}$ is normalized such that $\sum_i |x_i|^2 = 1$. This yields an exponentially compact representation of high-dimensional data~\cite{schuld2019quantum, larose2020robust}, but often demands deep and complex state preparation circuits.

After embedding, the quantum state is processed through a parameterized quantum circuit, commonly referred to as a variational quantum circuit (VQC). These circuits apply trainable unitary operations and are optimized using classical algorithms such as gradient descent or Adam to minimize task-specific loss functions, including mean squared error or cross-entropy~\cite{cerezo2021variational}. The expressibility of the variational ansatz, together with the loss landscape and optimizer choice, plays a critical role in convergence and final model performance~\cite{cerezo2021variational, mcclean2016theory}.

\subsection{Quantum Graph Attention Network}

\paragraph{Motivation and Overview.}
The Quantum Graph Attention Network (QGAT) integrates quantum circuits into the attention mechanism to enhance feature representation. Unlike classical models such as GAT and GATv2, which rely on linear or piecewise-linear transformations, QGAT leverages the nonlinear and high-dimensional mappings of quantum circuits induced by unitary evolution and entanglement. This enables richer feature interactions with fewer parameters. By embedding quantum layers into attention computation, QGAT achieves greater expressiveness without sacrificing scalability.

\paragraph{Quantum Data Embedding.}
QGAT adopts amplitude encoding to map node features into quantum states. Given a classical input vector $\mathbf{x} \in \mathbb{R}^d$, it is first normalized and padded to length $2^n$ (if necessary), and then encoded as:
\begin{equation}
|\psi(\mathbf{x})\rangle = \sum_{i=0}^{2^n - 1} x_i |i\rangle,
\end{equation}
where $n = \lceil \log_2 d \rceil$ is the number of qubits. This approach enables compact and expressive representation of high-dimensional node features using a logarithmic number of qubits.

Amplitude encoding is chosen for its ability to preserve global feature structure by directly mapping inputs to quantum state amplitudes. Compared to angle-based methods, it captures richer feature relationships with fewer gates and requires only logarithmic qubits relative to input size, reducing hardware demands while maintaining expressiveness.

\paragraph{Quantum Circuit Design.}
The quantum module in QGAT adopts a variant of variational quantum circuits based on strongly entangling layers~\cite{schuld2020circuit}. Each block begins with an amplitude-encoded input and applies parameterized single-qubit rotations followed by entangling operations.

Each single-qubit gate is implemented using a standard Z-Y-Z decomposition:
\begin{equation}
G_j = R_z(\mu_1) R_y(\mu_2) R_z(\mu_3),
\end{equation}
where $\mu_1$, $\mu_2$, and $\mu_3$ are trainable parameters, and $R_y(\theta)$, $R_z(\theta)$ denote Y- and Z-axis rotations, respectively. This decomposition is both universal and hardware-friendly, allowing expressive and stable control over state evolution with minimal circuit depth.

Entanglement is introduced using two-qubit gates placed after the single-qubit rotations. We use \textit{imprimitive} two-qubit operations capable of entangling initially separable states. A representative example is the controlled single-qubit gate:
\begin{equation}
C_a(G_b):|x\rangle |y\rangle \mapsto |x\rangle \otimes G_b^x |y\rangle,
\end{equation}
where $G$ is a single-qubit gate applied to qubit $b$ conditioned on the state of qubit $a$. In the special case $G = X$, this becomes a standard CNOT gate.

The qubit connectivity for entanglement is governed by a tunable range parameter, alternating across layers to allow both short- and long-range entanglement. A complete quantum block consists of a sequence of single-qubit rotations and controlled entangling gates:
\begin{equation}
B = \prod_{k=0}^{n-1} R^Y_k\, C_{ck}(P_{tk})\, \prod_{j=0}^{n-1} G_j,
\end{equation}
where $R^Y_k$ denotes a Y-rotation on qubit $k$, $C_{ck}(P_{tk})$ is a controlled phase gate, and $G_j$ is the Z-Y-Z rotation gate described above.

\begin{figure}[h]
    \centering
    \includegraphics[width=1.0\textwidth]{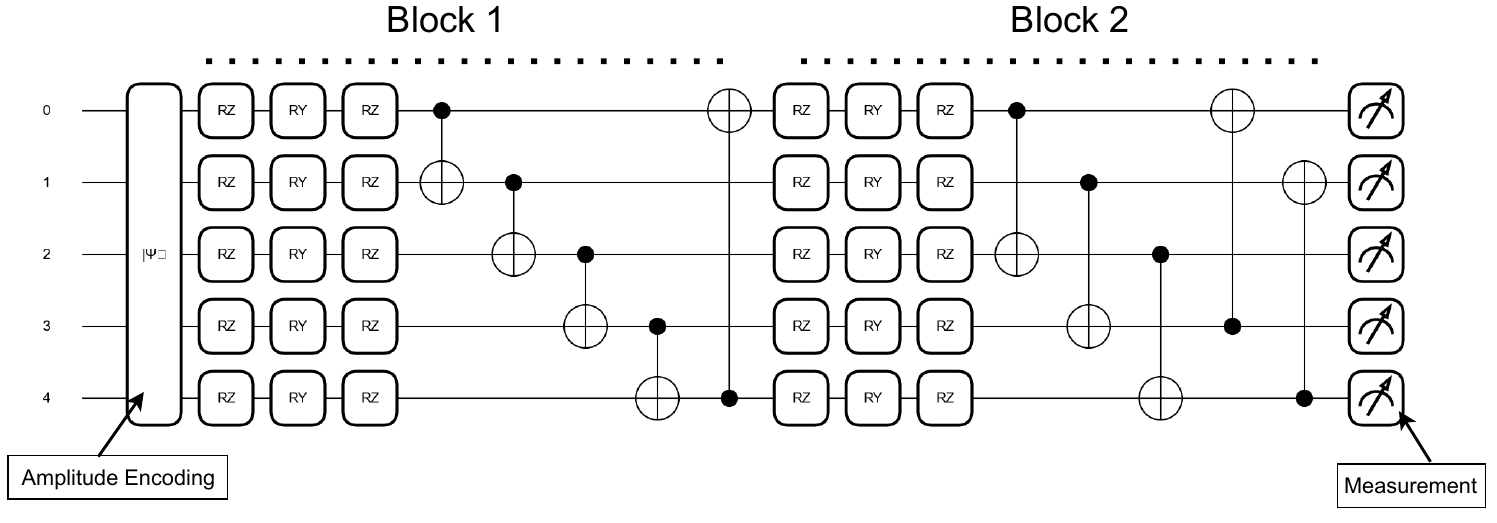}
    \caption{Example of a quantum circuit used in QGAT with $n=5$ qubits and two strongly entangling blocks. Each block includes parameterized single-qubit rotations ($R_y$, $R_z$) and entangling layers constructed using CNOT gates. After the final block, Pauli-$Z$ measurements are performed. The two-qubit entangling gates are CNOTs applied chronologically on qubits $i = 1,\dots,M$, where the target qubit is $(i + r) \bmod M$ and $r$ is a tunable \textit{range} parameter with $0 < r < M$. This configuration allows the circuit to achieve both local and long-range entanglement.}
    \label{fig:circuit}
\end{figure}

By stacking multiple such blocks, the circuit encodes classical features into high-dimensional entangled quantum states. This enhances the attention mechanism's capacity to model complex interactions. An example with five qubits and two entangling blocks is shown in Figure~\ref{fig:circuit}, illustrating how entanglement is distributed across the entire system.

\paragraph{Quantum Graph Attention Architecture.}
The proposed Quantum Graph Attention mechanism computes attention coefficients using a variational quantum circuit. Given input node features $\mathbf{h} \in \mathbb{R}^{N\times d}$, we first project them into multi-head representations via a shared linear transformation $\mathbf{W} \in \mathbb{R}^{d \times (h \cdot d_{\text{out}})}$. For each edge $(i,j)$, the projected features are concatenated:
\begin{equation}
\mathbf{a}_{ij} = [\mathbf{W}\mathbf{h}_i \| \mathbf{W}\mathbf{h}_j] \in \mathbb{R}^{2h d_{\text{out}}},
\end{equation}
where $[\cdot \| \cdot]$ denotes feature-wise concatenation.

To construct the quantum input, we further concatenate the original node features $\mathbf{h}_i$ and $\mathbf{h}_j$ with $\mathbf{a}_{ij}$ and apply a second projection $\mathbf{P}$:
\begin{equation}
\mathbf{a}'_{ij} = \mathbf{P}([\mathbf{a}_{ij} \| \mathbf{h}_i \| \mathbf{h}_j]) \in \mathbb{R}^{2^{n_q} \cdot \lceil h / n_q \rceil},
\end{equation}
where $n_q$ is the number of qubits in the quantum circuit.

To preserve essential structural and semantic information, we introduce a residual-enhanced design that concatenates the original node features with projected multi-head features. This enriches the quantum encoding and mitigates information loss from earlier transformations. The projection matrix $\mathbf{P}$ adjusts the input length to match the amplitude encoding dimension while retaining key feature correlations. The resulting vector $\mathbf{a}'_{ij}$ is then amplitude-encoded and processed by a quantum circuit. Although $\mathbf{P}$ compresses the input, the residual path helps retain fine-grained details. The circuit outputs multiple attention logits by measuring Pauli-$Z$ expectation values across qubits. If more heads are needed than available qubits, the circuit is executed repeatedly with the same input.

The quantum circuit applies the variational ansatz $U(\boldsymbol{\theta})$ to the amplitude-encoded state $|\psi(\mathbf{a}'_{ij})\rangle$. Each qubit $k$ produces one attention logit by measuring the expectation value of the Pauli-$Z$ observable:
\begin{equation}
e_{ij}^{(k)} = \langle \psi(\mathbf{a}'_{ij}) | U^\dagger(\boldsymbol{\theta}) Z_k U(\boldsymbol{\theta}) | \psi(\mathbf{a}'_{ij}) \rangle, \quad k = 1, \dots, n_q,
\end{equation}
where $Z_k$ is the Pauli-$Z$ operator acting on the $k$-th qubit.

When the number of required attention heads $h$ exceeds the available qubits $n_q$, the circuit is executed $n_{\mathrm{exec}}$ times:
\begin{equation}
n_{\mathrm{exec}} = \left\lceil \frac{h}{n_q} \right\rceil,
\end{equation}
and the outputs from all executions are concatenated to form the full set of $h$ attention logits.

These logits are normalized using softmax:
\begin{equation}
\alpha_{ij}^{(k)} = \frac{\exp(e_{ij}^{(k)})}{\sum\limits_{m \in \mathcal{N}_i} \exp(e_{im}^{(k)})},
\end{equation}
where $\mathcal{N}_i$ denotes the neighbors of node $i$, and $m$ iterates over all such neighbors in the normalization term.

No additional nonlinearities (e.g., LeakyReLU) are applied to $e_{ij}^{(k)}$, since the quantum circuit itself induces nonlinearity through its parameterized unitaries and measurement process.

The updated node representation is aggregated as follows:

\textbf{Concat:}
\begin{equation}
\mathbf{h}'i = \sigma\left( \mathrm{concat}{k=1}^{h} \sum_{j \in \mathcal{N}i} \alpha{ij}^{(k)} \mathbf{W}^{(k)} \mathbf{h}_j \right),
\end{equation}

\textbf{Mean:}
\begin{equation}
\mathbf{h}'i = \sigma\left( \frac{1}{h} \sum{k=1}^{h} \sum_{j \in \mathcal{N}i} \alpha{ij}^{(k)} \mathbf{W}^{(k)} \mathbf{h}_j \right),
\end{equation}
where $\mathbf{W}^{(k)}$ denotes the value projection matrix for the $k$-th head, and $\sigma$ is a non-linear activation function.

\begin{algorithm}[htbp]
\caption{Quantum Graph Attention Network (QGAT) Layer}
\label{alg:qgat}
\KwIn{Node features $\mathbf{h} \in \mathbb{R}^{N \times d}$, edge index $\mathcal{E}$, number of attention heads $h$, number of qubits $n_q$, number of quantum layers $L$, classical projection weights $\mathbf{W}$, compression matrix $\mathbf{P}$, activation $\sigma$, merge type (concat or mean).}
\KwOut{Updated node features $\mathbf{h}' \in \mathbb{R}^{N \times d'}$.}

Project node features into multi-head space:\
$\mathbf{h}'_i \gets \mathbf{W} \mathbf{h}_i$, for all $i$

\ForEach{edge $(i,j) \in \mathcal{E}$}{
    Concatenate projected and raw features:\
    $\mathbf{a}_{ij} \gets [\mathbf{h}'_i \| \mathbf{h}'_j \| \mathbf{h}_i \| \mathbf{h}_j]$

    Compress to match amplitude encoding input:\
    $\mathbf{a}'_{ij} \gets \mathbf{P}(\mathbf{a}_{ij}) \in \mathbb{R}^{2^{n_q} \cdot \lceil h / n_q \rceil}$

    Initialize quantum parameters $\boldsymbol{\theta}$

    \For{$m \gets 1$ \KwTo $\lceil h / n_q \rceil$}{
        Amplitude encode $\mathbf{a}'_{ij}$ as $|\psi(\mathbf{a}'_{ij})\rangle$ \\
        Apply variational circuit $U(\boldsymbol{\theta})$ with strongly entangling layers:\
        $|\psi'_{ij}\rangle \gets U(\boldsymbol{\theta}) |\psi(\mathbf{a}'_{ij})\rangle$ \\
        Measure Pauli-$Z$ expectation on all $n_q$ qubits:\ 
        $e_{ij}^{(k)} \gets \langle \psi(\mathbf{a}'_{ij}) | U^\dagger Z_k U | \psi(\mathbf{a}'_{ij}) \rangle$
    }
}

Normalize logits via softmax:\
$\alpha_{ij}^{(k)} \gets \frac{\exp(e_{ij}^{(k)})}{\sum\limits_{m \in \mathcal{N}_i} \exp(e_{im}^{(k)})}$

Update node embeddings:\\
\If{merge = concat}{
    $\mathbf{h}'_i \gets \sigma\left(\mathrm{concat}_{k=1}^{h}\sum_{j \in \mathcal{N}_i}\alpha_{ij}^{(k)}\mathbf{W}^{(k)}\mathbf{h}_j\right)$
}
\ElseIf{merge = mean}{
    $\mathbf{h}'_i \gets \sigma\left(\frac{1}{h}\sum_{k=1}^{h}\sum_{j \in \mathcal{N}_i}\alpha_{ij}^{(k)}\mathbf{W}^{(k)}\mathbf{h}_j\right)$
}

Apply residual connection and dropout\\
\Return updated node embeddings $\mathbf{h}'$
\end{algorithm}

\section{Evaluation}

We evaluate the performance of QGAT through a series of experiments aimed at assessing its robustness, expressiveness, and generalization across multiple graph learning tasks. Section~\ref{sec:noise} examines the model’s resilience to feature and structural perturbations. Section~\ref{sec:node} presents benchmark comparisons for node-level tasks, while Section~\ref{sec:link} evaluates link prediction performance. Across all experiments, we compare QGAT with established baselines, including GAT and GATv2, to highlight the benefits of quantum-enhanced attention. The evaluations cover both transductive and inductive settings to assess generalization to unseen nodes and graphs. Full training configurations—covering optimizer settings, learning rates, batch sizes, and early stopping criteria—are provided in Appendix~\ref{appendix:training_details}. All reported results reflect test set performance, evaluated at the best validation checkpoint over five independent runs.

\subsection{Robustness to Noise} \label{sec:noise}

We evaluate the robustness of QGAT on the \textit{ogbn-arxiv} dataset, a widely used citation graph benchmark with rich node features and semantic labels. This dataset is selected for its moderate size, structural complexity, and frequent use in evaluating graph learning models. We assess the model’s stability under two types of perturbations: feature noise and structural noise.

\paragraph{Feature Noise.}
To test robustness against feature corruption, we inject Gaussian noise into the node features and compare QGAT's performance to that of GAT and GATv2. Because QGAT uses amplitude encoding—which directly maps features into quantum state amplitudes—it is particularly important to assess its sensitivity to such distortions.

For a given noise level $\epsilon$, each input feature vector $\mathbf{x}$ is perturbed by adding zero-mean Gaussian noise:
\begin{equation}
\tilde{\mathbf{x}} = \mathbf{x} + \epsilon \cdot \mathcal{N}(0, \mathbf{I}),
\end{equation}
where $\epsilon \in {0.0, 0.01, 0.05, 0.1, 0.2}$ controls the magnitude of the perturbation. The same noise level is applied during both training and evaluation to ensure consistency.

\begin{figure}[htbp]
    \centering
    \begin{subfigure}[t]{0.48\textwidth}
        \centering
        \includegraphics[width=\textwidth]{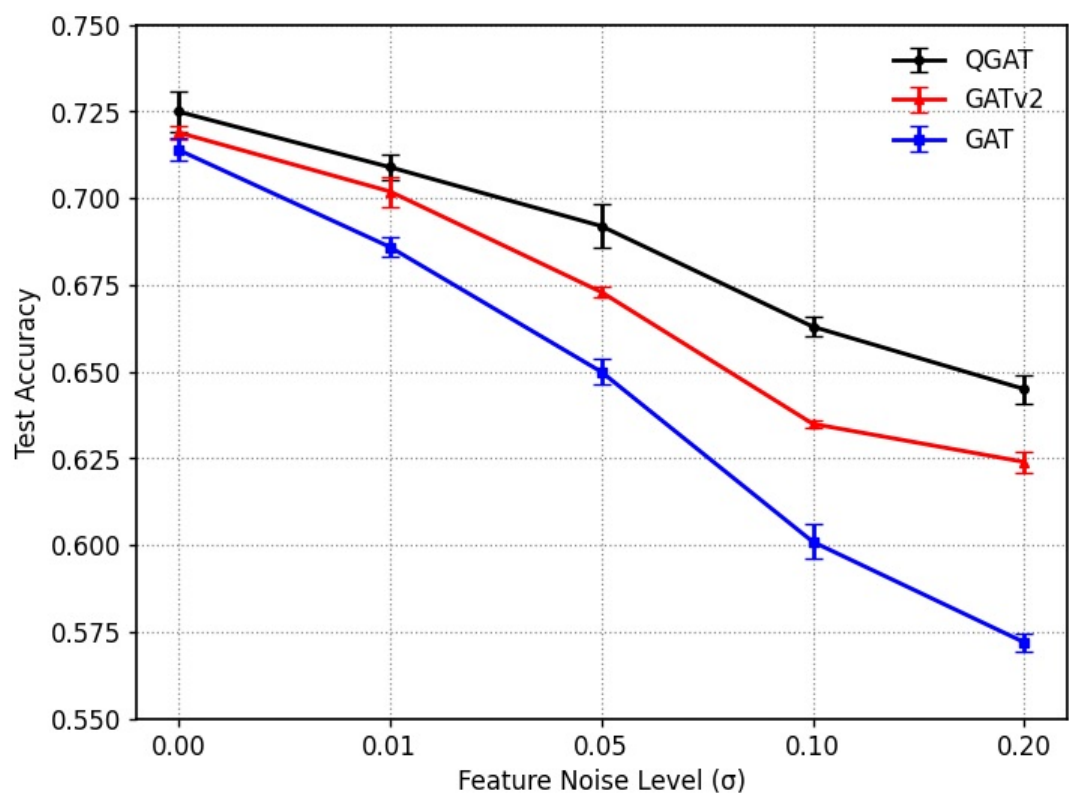}
        \caption{Feature noise. Test accuracy under Gaussian perturbation with $\sigma$.}
        \label{fig:fnoise}
    \end{subfigure}
    \hfill
    \begin{subfigure}[t]{0.48\textwidth}
        \centering
        \includegraphics[width=\textwidth]{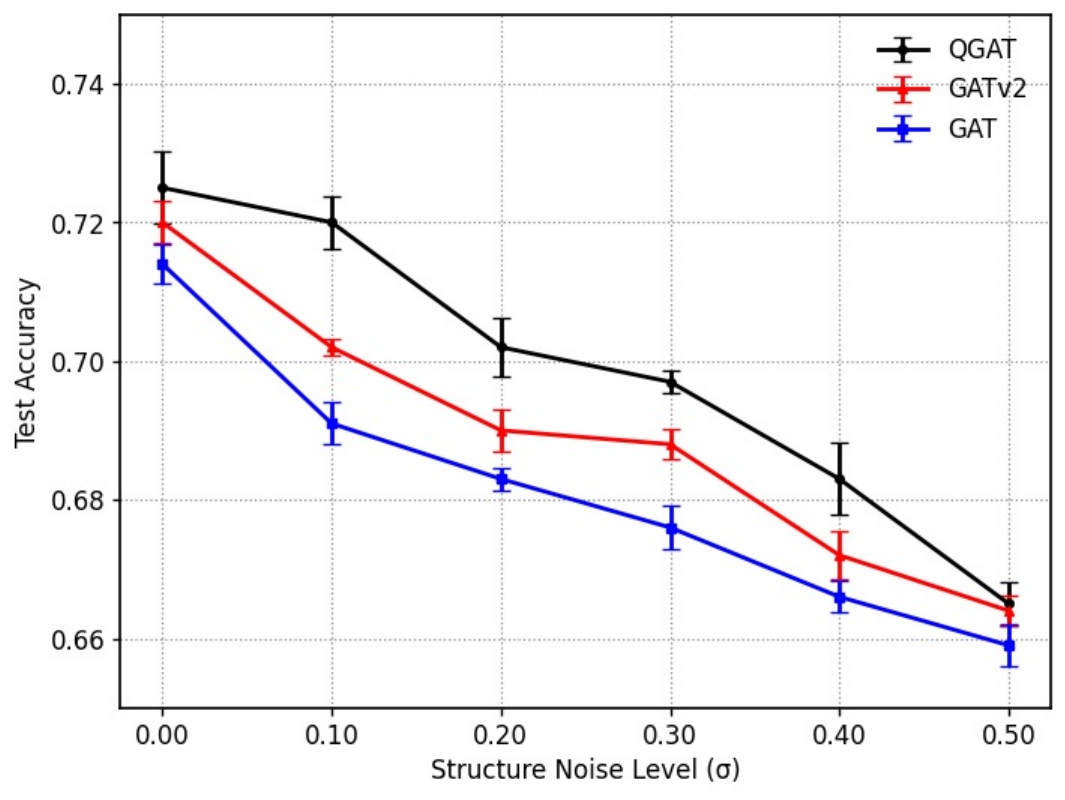}
        \caption{Structural noise. Test accuracy under random edge insertion with $\sigma$.}
        \label{fig:snoise}
    \end{subfigure}
    \caption{Robustness comparison of QGAT, GATv2, and GAT under feature and structural noise. Each point is averaged over 5 runs with standard deviation shown as error bars.}
    \label{fig:noise_all}
\end{figure}

Figure~\ref{fig:fnoise} shows test accuracy under increasing noise levels. While all models degrade with stronger perturbations, QGAT consistently outperforms GAT and GATv2 across all levels. This robustness is attributed to the global nature of amplitude embedding and the expressiveness introduced by quantum entanglement, which together help QGAT retain discriminative information even in noisy conditions.

\paragraph{Structural Noise.}
To assess QGAT’s robustness to topological perturbations, we introduce structural noise by randomly adding edges to the graph. We evaluate a range of predefined noise ratios $\eta \in {0.0, 0.1, 0.2, 0.3, 0.4, 0.5}$, where each value denotes the proportion of additional edges relative to the original edge count.

Given an undirected graph with $E$ edges and $N$ nodes, we sample $\lfloor \eta \cdot E \rfloor$ new edges by selecting random node pairs $(u, v)$ such that $u \neq v$ and $(u, v) \notin \mathcal{E}$. These randomly generated edges form a perturbation set $\mathcal{E}{\text{rand}}$, which is added to the original edge index:
\begin{equation}
\mathcal{E}{\text{noisy}} = \mathcal{E} \cup \mathcal{E}_{\text{rand}}.
\end{equation}

Figure~\ref{fig:snoise} presents the test accuracy under increasing structural noise levels. As noise increases, the performance of all models declines. However, QGAT consistently outperforms both GAT and GATv2, maintaining a higher accuracy margin especially under moderate to severe noise. This result suggests that QGAT is more resilient to irregular or corrupted graph topologies, likely due to the expressive relational modeling induced by amplitude encoding and quantum entanglement.

\subsection{Node Property Prediction} \label{sec:node}

We evaluate the effectiveness of QGAT in node property prediction tasks under two settings: \textit{transductive}, where the entire graph is available during training, and \textit{inductive}, where the model must generalize to unseen nodes or subgraphs. These settings offer complementary insights into the model's ability to capture both local neighborhood patterns and broader structural dependencies.

\paragraph{Transductive}

We conduct experiments on three widely used transductive node classification benchmarks: \textit{Pubmed}, \textit{ogbn-arxiv}, and \textit{ogbn-products}. These datasets test the model’s capacity to learn from full graph structures, with access to all nodes and edges during training. QGAT is evaluated under the same experimental settings as classical baselines, including GAT and GATv2. As shown in Table~\ref{tab:transductive}, QGAT consistently achieves higher accuracy across all datasets, demonstrating its superior ability to capture structural and semantic node-level relationships.

\begin{table}[h]
\centering
\caption{Accuracy (\%) on transductive node classification benchmarks. All results are averaged over 5 runs; standard deviations are reported.}
\label{tab:transductive}
\begin{tabular}{lccc}
\toprule
\textbf{Model} & \textbf{Pubmed} & \textbf{ogbn-arxiv} & \textbf{ogbn-products} \\
\midrule
GAT \cite{brody2021attentive}     & $78.1 \pm 0.59$ & $71.54 \pm 0.3$ & $79.04 \pm 1.54$ \\
GATv2 \cite{brody2021attentive}  & $78.5 \pm 0.38$ & $71.87 \pm 0.25$ & $80.63 \pm 0.7$ \\
QGAT    & $\mathbf{79.2 \pm 0.62}$ & $\mathbf{73.62 \pm 0.42}$ & $\mathbf{82.10 \pm 2.31}$ \\

\bottomrule
\end{tabular}
\end{table}

Based on the results in Table~\ref{tab:transductive}, QGAT achieves the highest accuracy across all three transductive node classification benchmarks. It surpasses GATv2 by approximately $+0.7\%$ on \textit{Pubmed}, $+1.75\%$ on \textit{ogbn-arxiv}, and $+1.47\%$ on \textit{ogbn-products}. These consistent improvements, accompanied by low standard deviations, indicate that QGAT offers a more expressive and stable mechanism for capturing node-level dependencies when the full graph is available during training.

\paragraph{Inductive.}

To evaluate the generalization ability of QGAT under inductive settings, we conduct node classification benchmarks on two widely used datasets: \textit{PPI} and \textit{ogbn-proteins}. These datasets involve either multiple disjoint graphs (PPI) or large-scale protein interaction networks (ogbn-proteins), where test nodes or subgraphs are inaccessible during training. This setup simulates realistic scenarios where models must transfer learned representations to previously unseen structures. As summarized in Table~\ref{tab:inductive}, QGAT is evaluated alongside GAT and GATv2 under identical configurations.

\begin{table}[htbp]
\centering
\caption{Inductive node classification performance on PPI and ogbn-proteins. Metric: Micro-F1 (PPI) and ROC-AUC (ogbn-proteins). All results are averaged over 5 runs; standard deviations are reported.}
\label{tab:inductive}
\begin{tabular}{lcc}
\toprule
\textbf{Model} & \textbf{PPI (Micro-F1)} & \textbf{ogbn-proteins (ROC-AUC)} \\
\midrule
GAT \cite{hamilton2017inductive}     & $97.3 \pm 0.20$ & $78.63 \pm 1.62$ \\
GATv2   & $98.2 \pm 0.25$ & $\mathbf{79.52 \pm 0.55}$ \cite{brody2021attentive} \\
QGAT    & $\mathbf{98.9 \pm 0.12}$ & $79.41 \pm 0.21$ \\
\bottomrule
\end{tabular}
\end{table}

QGAT demonstrates strong inductive generalization across both datasets. On PPI, it outperforms both GAT and GATv2, achieving a Micro-F1 score of $98.9\%$, reflecting its ability to transfer representations across disconnected graphs. On ogbn-proteins, QGAT attains a ROC-AUC of $79.41\%$, performing competitively with GATv2 while exhibiting lower variance, indicating more stable prediction across protein interaction structures.

Since prior literature does not report GATv2 results on the PPI dataset, we implemented a three-layer GATv2Conv~\cite{brody2021attentive} baseline using attention heads $[8, 8, 4]$, hidden dimension of $256$, dropout rate of $0.5$, and LayerNorm applied after each layer.

\subsection{Link Prediction} \label{sec:link}

To assess QGAT’s ability to capture implicit relationships between node pairs, we evaluate its performance on link prediction tasks using two representative datasets: \textit{ogbl-collab} and \textit{ogbl-citation2}. These datasets span diverse domains including collaboration networks, drug-drug interactions, and citation graphs. The model is trained to predict the existence of edges based on learned node representations, providing insight into its capacity for structural reasoning and generalization.

\begin{table}[htbp]
\centering
\caption{Link prediction performance on OGB benchmarks. Metric: Hits@50 (\%) for \textit{ogbl-collab}, and Mean Reciprocal Rank (MRR) for \textit{ogbl-citation2}. All results are averaged over 5 runs; standard deviations are reported.
}
\label{tab:link_prediction}
\begin{tabular}{lccc}
\toprule
\textbf{Model} & \textbf{ogbl-collab}  & \textbf{ogbl-citation2} \\
\midrule
GAT \cite{brody2021attentive}     & $46.63 \pm 2.80$ & $75.95 \pm 1.31$ \\
GATv2 \cite{brody2021attentive}   & $49.7 \pm 3.08$ & $80.14 \pm 0.71$ \\
QGAT    & $\mathbf{51.2 \pm 1.92}$ & $\mathbf{82.2 \pm 1.27}$ \\
\bottomrule
\end{tabular}
\end{table}

As shown in Table~\ref{tab:link_prediction}, QGAT achieves the highest Hits@50 on \textit{ogbl-collab} with $51.2\%$, outperforming GAT and GATv2 by $4.57\%$ and $1.5\%$, respectively. On the more challenging \textit{ogbl-citation2} benchmark, QGAT attains the top mean reciprocal rank (MRR) of $82.2\%$, surpassing both baselines. These results highlight QGAT’s enhanced ability to model latent structural dependencies, driven by the nonlinear and entangled transformations introduced through the quantum attention mechanism.

\section{Conclusion}
We introduced the Quantum Graph Attention Network (QGAT), a hybrid architecture integrating variational quantum circuits into graph neural network attention mechanisms. By using amplitude encoding and strongly entangled quantum layers, QGAT captures rich nonlinear interactions and efficiently implements multi-head attention within a single quantum circuit. This design reduces parameter overhead and enhances representational capacity, enabling compact yet expressive models. Experiments across various benchmarks—including transductive and inductive node classification, as well as link prediction—show that QGAT consistently outperforms classical baselines such as GAT and GATv2. Notably, these performance gains persist even under strict hyperparameter constraints due to memory limitations, underscoring the model's efficiency and robust generalization. Moreover, evaluations under noisy conditions highlight the increased robustness of amplitude-encoded quantum representations against feature-level and structural perturbations. Thus, QGAT effectively demonstrates the practicality and benefits of quantum-enhanced attention in realistic graph learning tasks.

\paragraph{Discussion and Future Work.}
Our experiments demonstrate that QGAT excels on complex, multi-label datasets such as PPI, highlighting its effectiveness in capturing intricate relational structures found in biological interaction networks and molecular systems. Additionally, QGAT shows competitive results on link prediction tasks, reflecting its capacity to learn fine-grained edge-level relationships. Thanks to its modular design, QGAT can directly replace classical GAT or GATv2 layers, facilitating easy integration into various existing architectures and training workflows.

However, hardware scalability remains a key limitation. Our current implementation does not support distributed quantum-classical execution across multiple GPUs or nodes, restricting hyperparameter configurations and limiting applicability to large-scale graphs. Addressing this constraint—through circuit batching, tensor parallelism, or quantum-aware distributed scheduling—is essential for scaling QGAT to larger and more computationally intensive graph tasks in the future.



\appendix
\section{Training Details}
\label{appendix:training_details}

QGAT is implemented with \texttt{Pennylane}\cite{bergholm2022pennylane} for quantum circuit simulation and \texttt{PyTorch} for classical computation. All experiments are conducted on a single NVIDIA H100 GPU.

The quantum module introduces additional memory overhead due to circuit simulation and multi-head execution. As a result, we restrict the hyperparameter search space and do not explore larger hidden dimensions or wider attention heads. Despite this limitation, QGAT consistently achieves competitive performance across all benchmarks.

\subsection{Robustness to Noise} \label{sec:Robustness to Noise}

For QGAT, we perform a grid search over the following hyperparameters: learning rate $\in \{1\times10^{-2}, 2\times10^{-3}, 1\times10^{-3}, 1\times10^{-4}\}$, hidden dimension $\in \{32, 64, 128\}$, and dropout rate $\in \{0.4, 0.5, 0.6\}$, and strong entanglement layer $\in \{2, 3, 4\}$. The model uses three graph attention layers with attention heads configured as $[4, 4, 4]$ or $[6, 6, 4]$, depending on the dataset. Optimization is performed using the AdamW optimizer along with a CosineAnnealingLR scheduler. All training is carried out using mixed-precision (AMP) to reduce memory usage and accelerate convergence.

Experiments on the \textit{ogbn-arxiv} dataset are conducted using the GraphSAINT mini-batch sampling strategy~\cite{zeng2020graphsaint} to reduce memory consumption and enable scalable training. This sampling approach allows QGAT to operate effectively on large graphs without exceeding GPU memory limits.

For baseline models (GAT and GATv2), we adopt the hyperparameter configurations reported in the original papers~\cite{velivckovic2017graph, brody2021attentive} to ensure fair and consistent comparisons.

\subsection{Node Property Prediction}

\paragraph{Transductive}
We evaluate QGAT on three node classification benchmarks: \textit{Pubmed}, \textit{ogbn-arxiv}, and \textit{ogbn-products}, each with dataset-specific hyperparameter configurations. Across all datasets, training is performed using the AdamW optimizer with a CosineAnnealingLR scheduler. Mixed-precision training (AMP) is employed to reduce GPU memory usage and improve efficiency.

For the \textit{Pubmed} dataset, we perform a grid search over the following hyperparameters: learning rate $\in \{1\times10^{-2}, 2\times10^{-3}, 1\times10^{-3}, 1\times10^{-4}\}$, hidden dimension $\in \{8, 16, 32\}$, and dropout $\in \{0.3, 0.4, 0.5, 0.6\}$, and strong entanglement layer $\in \{2, 3, 4\}$. The architecture uses two QGAT layers, with attention head settings including $[6, 4]$, $[4, 1]$, and $[4, 4]$.

For the \textit{ogbn-arxiv} dataset, we adopt the same hyperparameter configurations as in the robustness experiments: learning rate $\in \{1\times10^{-2}, 2\times10^{-3}, 1\times10^{-3}, 1\times10^{-4}\}$, hidden dimension $\in \{32, 64, 128\}$, dropout $\in \{0.4, 0.5, 0.6\}$, three layers, and attention head configurations $[4, 4, 4]$ or $[6, 6, 4]$, and strong entanglement layer $\in \{2, 3, 4\}$. Training on ogbn-arxiv is performed with the GraphSAINT sampling strategy to enable scalable training on the full graph.

For the \textit{ogbn-products} dataset, we use a similar setup to ogbn-arxiv, with a grid over learning rate $\in \{1\times10^{-2}, 2\times10^{-3}, 1\times10^{-3}, 1\times10^{-4}\}$, hidden dimension $\in \{32, 64, 128\}$, dropout $\in \{0.4, 0.5, 0.6\}$, and attention heads $[4, 4, 4]$ or $[6, 6, 4]$ across three QGAT layers, and strong entanglement layer $\in \{2, 3, 4\}$. Due to the scale of the dataset, GraphSAINT is also used here to reduce computational and memory overhead.

\paragraph{Inductive.}

For the \textit{PPI} dataset, we perform a grid search over learning rate $\in \{1\times10^{-2}, 2\times10^{-3}, 1\times10^{-3}, 1\times10^{-4}\}$, hidden dimension $\in \{32, 64, 128, 256\}$, and dropout $\in \{0.0, 0.3, 0.5\}$, and strong entanglement layer $\in \{2, 3, 4\}$. The model consists of three QGAT layers with attention head configurations $[6, 6, 4]$ and $[8, 8, 4]$. Optimization is carried out using AdamW with a CosineAnnealingLR scheduler. All training is performed using mixed-precision training (AMP), and a fixed batch size of 2 is used for training, validation, and testing to accommodate GPU memory constraints.

For the \textit{ogbn-proteins} dataset, we use a similar setup, with learning rate $\in \{1\times10^{-2}, 2\times10^{-3}, 1\times10^{-3}, 1\times10^{-4}\}$, hidden dimension $\in \{32, 64, 128\}$, dropout $\in \{0.0, 0.3, 0.5\}$, three QGAT layers using either $[6, 6, 4]$ or $[8, 8, 4]$ heads, and strong entanglement layer $\in \{2, 3, 4\}$. GraphSAINT is applied for scalable mini-batch training due to the dataset's large size. The model is optimized with AdamW and a CosineAnnealingLR scheduler, using mixed-precision training to reduce memory usage and improve runtime efficiency.

\subsection{Link Prediction}

For the \textit{ogbl-collab} dataset, we perform a grid search over learning rate $\in \{1\times10^{-2}, 2\times10^{-3}, 1\times10^{-3}, 1\times10^{-4}\}$, hidden dimension $\in \{64, 128, 256\}$, and dropout rate $\in \{0.3, 0.4, 0.5, 0.6\}$, and strong entanglement layer $\in \{2, 3, 4\}$. The model is composed of three QGAT layers with attention heads configured as $[4, 4, 4]$ or $[6, 6, 4]$. To handle the large scale of the dataset, we apply GraphSAINT~\cite{zeng2020graphsaint} for sampling-based training. Optimization is done using AdamW with a CosineAnnealingLR scheduler, and mixed-precision training is used throughout.

For the \textit{ogbl-citation2} dataset, we adopt a similar setup with adjusted learning rate grid $\in \{2\times10^{-3}, 1\times10^{-3}, 1\times10^{-4}, 5\times10^{-5}\}$, while using the same hidden dimension range $\{64, 128, 256\}$, dropout values $\{0.3, 0.4, 0.5, 0.6\}$, strong entanglement layer $\in \{2, 3, 4\}$, and three-layer QGAT architecture. Attention heads are also set to $[4, 4, 4]$ or $[6, 6, 4]$, and GraphSAINT is employed to enable efficient training. The optimizer and scheduler choices remain consistent with previous settings.

\section{Dataset details}

\subsection{Node Property Prediction}
This section summarizes the datasets used for node property prediction tasks, including both transductive and inductive settings. Table~\ref{tab:node_datasets} reports key statistics such as the number of nodes, edges, feature dimensions, and task type. These datasets span various domains and settings, providing a diverse benchmark to evaluate the generalization and robustness of QGAT. In particular, \textit{Pubmed}, \textit{ogbn-arxiv}, and \textit{ogbn-products} are evaluated under transductive learning, while \textit{PPI} and \textit{ogbn-proteins} are used in inductive scenarios.

\begin{table}[htbp]
\centering
\caption{Statistics and settings of datasets used for node property prediction.}
\label{tab:node_datasets}
\begin{tabular}{lcccc}
\toprule
\textbf{Dataset} & \textbf{\#Nodes} & \textbf{\#Edges} & \textbf{\#Features} & \textbf{Task} \\
\midrule
Pubmed & 19,717 & 44,338 & 500 & Node classification  \\
ogbn-arxiv & 169,343 & 1,166,243 & 128 & Node classification  \\
ogbn-products & 2,449,029 & 61,859,140 & 100 & Node classification  \\
PPI & 56,944 & 818,716 & 50 & Node classification (multi-label)  \\
ogbn-proteins & 132,534 & 39,561,252 & 8 (edge features) & Node classification (binary) \\
\bottomrule
\end{tabular}
\end{table}

\subsection{Link Prediction}

We use two benchmark datasets from the Open Graph Benchmark (OGB) suite to evaluate the link prediction capability of QGAT. These datasets span different domains—collaboration networks and citation graphs—challenging the model to capture implicit structural relationships between node pairs. Table~\ref{tab:link_datasets} summarizes the key statistics and task definitions.

\begin{table}[htbp]
\centering
\caption{Statistics and settings of datasets used for link prediction.}
\label{tab:link_datasets}
\begin{tabular}{lcccc}
\toprule
\textbf{Dataset} & \textbf{\#Nodes} & \textbf{\#Edges} & \textbf{\#Features} & \textbf{Task} \\
\midrule
ogbl-collab & 235,868 & 1,285,465 & 128 & Link prediction (Hits@50) \\
ogbl-citation2 & 2,927,963 & 30,000,000 & 128 & Link prediction (MRR) \\
\bottomrule
\end{tabular}
\end{table}

\section{Parameter Comparison}

\begin{table}[htbp]
\centering
\caption{Parameter comparison of GAT, GATv2, and QGAT on the PPI dataset.}
\label{tab:param_compare}
\begin{tabular}{lcccc}
\toprule
Model & Hidden Dim & Layers & Attention Heads & Parameters \\\midrule
GAT & 256 & 3 & [8, 8, 4] & 6,439,545 \\
GATv2 & 256 & 3 & [8, 8, 4] & 12,838,521 \\
QGAT (\textit{\scriptsize Strong Entanglement Layer = 2}) & 256 & 3 & [8, 8, 4] & 6,720,017 \\
QGAT (\textit{\scriptsize Strong Entanglement Layer = 3}) & 256 & 3 & [8, 8, 4] & 6,720,053 \\
QGAT (\textit{\scriptsize Strong Entanglement Layer = 4}) & 256 & 3 & [8, 8, 4] & 6,720,089 \\
\bottomrule
\end{tabular}
\end{table}

Table~\ref{tab:param_compare} shows that QGAT, despite integrating quantum circuits and employing strong entanglement, has a parameter count only slightly higher than GAT and significantly lower than GATv2. This indicates that QGAT effectively enhances nonlinear expressivity and feature interactions with minimal additional parameters, achieving a balance between performance and parameter efficiency.

However, due to the use of a PennyLane quantum simulator combined with PyTorch in hybrid mode, training efficiency has not been fully optimized. As a result, the training time of QGAT is approximately five to six times longer than that of GAT and GATv2 under the same hardware conditions. 



\end{document}